\documentclass{article}
\usepackage{ijcai18}

\usepackage{times}
\usepackage{xcolor}
\usepackage{soul}
\usepackage{amssymb}
\usepackage[utf8]{inputenc}
\usepackage[small]{caption}

\usepackage{helvet}  
\usepackage{courier}  
\usepackage{url}  
\usepackage{graphicx}  
\usepackage{amsmath,bm,amsfonts}
\usepackage{multirow,subfigure}
\usepackage{color}
\usepackage{float}

\DeclareMathOperator*{\argmin}{arg\,min}

\newcommand{\xv}[0]{{{\bf x}}}
\newcommand{\yv}[0]{{{\bf y}}}

\newcommand{\muv}[0]{{{\bm \mu}}}
\newcommand{\Sigmav}[0]{{{\bf \Sigma}}}

\newcommand{\Xv}[0]{{{\bf X}}}
\newcommand{\Yv}[0]{{{\bf Y}}}

\newcommand{\Dv}[0]{{{\bf D}}}
\newcommand{\mc}[1]{{{\mathcal{#1}}}}

\title{Adversarial Constraint Learning for Structured Prediction}

\author{Hongyu Ren, Russell Stewart, Jiaming Song, Volodymyr Kuleshov, Stefano Ermon\\
Department of Computer Science, Stanford University\\
\{hyren, stewartr, tsong, kuleshov, ermon\}@cs.stanford.edu}

\begin{document}

\maketitle

\begin{abstract}
Constraint-based learning 
reduces the burden of collecting labels by having users specify general properties of structured outputs, such as constraints imposed by physical laws.
We propose a novel framework for simultaneously learning these constraints and using them for supervision, bypassing the difficulty of using domain expertise to manually specify constraints. 
Learning requires a black-box simulator of structured outputs, which generates valid labels, but need not model their corresponding inputs or the input-label relationship. 
At training time, we constrain the model to produce outputs that cannot be distinguished from simulated labels by adversarial training.
Providing our framework with a small number of labeled inputs gives rise to a new semi-supervised structured prediction model; 
we evaluate this model on multiple tasks ---
tracking, pose estimation and time series prediction --- and find that
it achieves high accuracy with only a small number of labeled inputs. In some cases, no labels are required at all. 
\end{abstract}

\section{Introduction}
Large labeled datasets are a key component for building state-of-the-art systems in many applications of machine learning, including image recognition,  
machine translation,  
and speech recognition. 
Collecting such datasets can be expensive, which has driven significant research interest in unsupervised, semi-supervised, and weakly supervised learning approaches \cite{radford2015unsupervised,kingma2014semi,papandreou2015weakly,ratner2016data}.

Constraint-based learning is a recently proposed form of weak supervision which aims to reduce the need for labeled inputs
by having users supervise algorithms through general properties that hold over the label space \cite{shcherbatyi2016convexification,stewart2017label}.
Examples of such properties include logical rules \cite{richardson2006markov,chang2007guiding,choi2015tractable,xu2017semantic} or physical laws \cite{stewart2017label,ermon2015pattern}.

Unlike labels --- which only apply to their corresponding inputs ---  properties used in a constraint-based learning approach are specified once for the entire dataset, providing an opportunity for more cost-efficient supervision.
Algorithms supervised with explicit constraints have shown promising results 
in 
object detection \cite{stewart2017label}, preference learning \cite{choi2015tractable}, materials science~\cite{ermon2012smt}, and semantic segmentation \cite{pathak2015constrained}.

However, describing the high level invariants of a dataset 
may also require a non-trivial amount of effort.
First, designing constraints requires strong domain expertise.
Second, in the case of high dimensional labels, it is difficult to encode the constraints using simple formulas.  
For example, suppose we want to constrain a pedestrian joint detector to produce skeletons that ``look like a walking person''; in this case, it is difficult to capture invariants over human poses with simple logical or algebraic formulas that an annotator could specify.
Third, constraints may change over time and across tasks; designing new constraints for new tasks may not scale in many practical applications. 

\begin{figure}[t]
\centering
\includegraphics[width=0.45\textwidth]{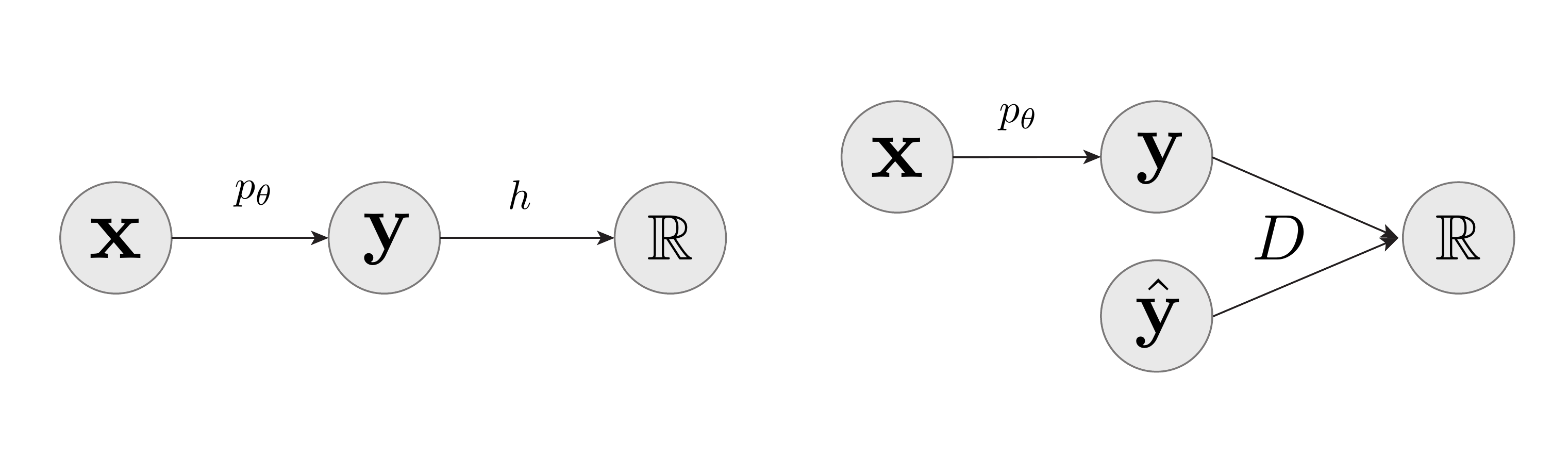}
\caption{Constraint learning allows us to learn a conditional probabilistic model $p_\theta(\yv|\xv)$ (parameterized by $\theta$)  
without direct labels by specifying properties $h$ that holds over the output space. In prior work (left), $h$ is defined as a formula describing known invariants. In this paper (right), we propose to instead learn $h$ through an auxiliary classifier $D_\phi$ (parameterized by $\phi$) that discriminates $\yv$ (provided by $p_\theta(\yv|\xv)$) from $\hat{\yv}$ (provided by an additional source unrelated to $\xv$, such as a simulator).}
\label{fig:concept}
\end{figure}

In this paper, we propose an \textit{implicit} approach to constraint learning, in which invariants are automatically learned from a small set of representative label samples (see Figure \ref{fig:concept}).\footnote{Please find source code in \url{https://github.com/hyren/acl}}
These samples do not need to be tied to corresponding inputs (as in supervised learning) and may come from a black-box simulator that abstracts away physics-based formulas or produces examples of labels collected by humans. Such simulators include physics engines, humanoid simulators from robotics, or driving simulators~\cite{li2017infogail}.

Inspired by recent advances in implicit (likelihood-free) generative modeling, we capture the distribution of outputs using an approach based on adversarial learning \cite{goodfellow2014generative}. 
Specifically, we train two distinct learners: a primary model for the task at hand and an auxiliary classification algorithm called discriminator. During training, we constrain the main model such that its outputs cannot be distinguished by the discriminator from representative (true) label samples, thus forcing it to capture the structure of the label space. This approach forms a novel adversarial framework for performing weak supervision with learned constraints, which we call adversarial constraint learning.

Although constraint learning does not require input-label pairs, providing such pairs can improve performance and turns our problem into an instance of semi-supervised learning. 
In this setting, our approach combines supervised learning on a small labeled dataset with constraint learning on a large unlabeled set, where constraint learning enforces that the structure of predictions on unlabeled data matches the structure observed in the labeled data.
Experimental results demonstrate that this method performs better than state-of-the-art semi-supervised learning methods on a variety of structured prediction problems.

\section{Background}
In this section, we introduce structured prediction and constraint-based learning. The next section will expand upon these subjects to introduce the proposed adversarial constraint learning framework. 

\subsection{Structured Prediction}
Our work focuses on structured prediction, a form of supervised learning, in which the outputs $\yv \in \mathcal{Y}$ can be a complex object such as a vector, a tree, or a graph \cite{koller2009probabilistic}. 
We capture the distribution of $\yv$ using a conditional probabilistic model $p_\theta(\yv|\xv)$ parameterized by $\theta \in \Theta$. 
A model $p_\theta(\yv|\xv)$ maps each input $\xv \in \mathcal{X}$ to the corresponding output distribution $p_\theta(\yv) \in \mathcal{P}(\mathcal{Y})$, where $\mathcal{P}(\mathcal{Y})$ denotes all the probability distributions over $\mathcal{Y}$. 
For example, we may take $p_\theta(\yv|\xv)$ to be a Gaussian distribution $\mathcal{N}(\muv_\theta(\xv),\Sigmav_\theta(\xv))$ with mean $\muv_\theta(\xv)$ and variance $\Sigmav_\theta(\xv)$.

A standard approach to learning $p_\theta(\yv|\xv)$ (or $p_\theta$ as an abbreviation) is to solve an optimization problem of the form
\begin{equation}
\label{eq:supervise}
\theta^{\ast}=\argmin_{\theta \in {\Theta}}\sum_{i=1}^{n} \ell (p_\theta(\yv|\xv_i),\yv_i) + R(p_\theta)
\end{equation}
over a labeled dataset $\Dv = \{(\xv_1,\yv_1), \cdots, (\xv_n, \yv_n)\}$. 
A typical supervised learning objective is comprised of a loss function $\ell: \mathcal{P}(\mathcal{Y})\times{\mathcal{Y}}\to\mathbb{R}$  
and a regularization term $R: \mathcal{P}(\mathcal{Y}) \to \mathbb{R}$ that encourages non-degenerate solutions or solutions that incorporate prior knowledge \cite{stewart2017label}. 

\subsection{Constraint-Based Learning}
Collecting a large labeled dataset for supervised learning can often be tedious.
Constraint-based learning is a form of weak supervision which instead asks users to specify high-level constraints over the output space, such as logical rules or physical laws \cite{shcherbatyi2016convexification,stewart2017label,richardson2006markov,xu2017semantic}. For example, in an object tracking task where $\mathcal{Y}$ corresponds to the space of joint positions over time, we expect correct outputs to be consistent with the laws of physical mechanics.

Let $\Xv = \{\xv_{1}, \cdots, \xv_m\}$ be an unlabeled dataset of inputs.
Formally, constraints can be specified via a function $h: \mathcal{P}(\mathcal{Y})\to\mathbb{R}$, which penalizes conditional probabilistic models $p_\theta(\yv|\xv)$ that are inconsistent with known high-level structure of the label space. 
Learning from constraints proceeds by optimizing the following objective:
\begin{equation}
\label{eq:CLobejctive}
\hat{\theta}^{\ast}=\argmin_{\theta \in {\Theta}}\sum_{i=1}^{m}h(p_\theta(\yv|\xv_i))+R(p_\theta)
\end{equation}
over $\Xv$. By solving this optimization problem, we look for a probabilistic model parameterized by $\hat{\theta}^{\ast}$ that satisfies known constraints when applied to the unlabeled dataset $\Xv$ (through the $h$ term), and is likely a priori (through the $R(p_\theta)$ term). Note that although the constraint $h$ 
is data-dependent, \emph{it does not require explicit labels}. For example, in object tracking we could ask that when making predictions on $\Xv$, joint positions over time are consistent with known kinematic equations, with $h$ measuring how the output distribution from $p_\theta$ deviates from those equations. 
The regularization term can be used to avoid overly complex and/or degenerate solutions, and may include $L1$, $L2$, or entropy regularization terms. 
Stewart and Ermon~\cite{stewart2017label} have shown that a model learned with the objective described in Eq. \ref{eq:CLobejctive} can learn to track objects.

\section{Adversarial Constraint Learning}
The process of manually specifying high level constraints, $h$, can be time-consuming and may require significant domain expertise. Such is the case in pose estimation, where it is difficult to  describe high dimensional rules for joints movements precisely; but the large availability of unpaired videos and motion capture data makes constraint learning attractive in spite of the difficulty of providing high dimensional constraints.

In the sciences, discovering general invariants is often a data-driven approach; for example, the laws of physics are often discovered by validating hypotheses with experimental results. 
Motivated by this, we propose in this section a novel framework for learning constraints from data. 

\subsection{Learning Constraints from Data}

Suppose we have a dataset of inputs $\Xv=\{\xv_1,\dots,\xv_m\}$, 
a dataset of labels $\Yv=\{\yv_1,\dots,\yv_k\}$, 
and a set $\Dv=\{(\xv_1,\yv_1),\dots,(\xv_n,\yv_n)\}$ that describes correspondence between some elements of $\Xv$ and $\Yv$. We denote the empirical distributions of $\Xv$, $\Yv$ and $\Dv$ as $p(\xv)$, $p(\yv)$ and $p(\xv, \yv)$ respectively.
Note that $\Yv$ can come from either a simulator (such as one based on physical rules), or from some other source of data (such as motion captures of people for which we have no corresponding videos).

Let us first consider the setting where $\Dv = \varnothing$; i.e. there are inputs and labels but no correspondence between them. In spite of the lack of correspondences, we will see that constraints $h$ can be learned from the prior knowledge that the same underlying distribution generates both the empirical labels $\Yv$ and the structured predictions obtained from applying our model to $\Xv$. These learned constraints can then be used for supervision. Let structured predictions be given by the following $\textit{implicit}$ sampling procedure:
\begin{align}
\xv \sim p(\xv) \quad, \quad \yv \sim p_\theta(\yv|\xv)
\end{align} 
where 
$p_\theta(\yv|\xv)$ is a (parameterized) conditional distribution of outputs given inputs.
Discarding $\xv$, the above procedure corresponds to sampling from the marginal distribution over $\mathcal{Y}$, $p_\theta(\yv) = \int p_\theta(\yv|\xv)p(\xv) \mathrm{d}\xv$. 

Labels drawn from $p(\yv)$ should have high likelihood values in $p_\theta(\yv)$, but optimizing this objective directly is computationally infeasible; evaluating the marginal likelihood $p_\theta(\yv)$ exactly is expensive due to the integration over $p(\xv)$. 
Instead, we formulate the task of learning a constraint loss $h$ from $p(\yv)$ through a likelihood-free approach using the framework of generative adversarial learning~\cite{goodfellow2014generative}, which only requires samples from $p_\theta(\yv)$ and $p(\yv)$.

We introduce an auxiliary classifier $D_\phi$ (parametrized by $\phi$) called discriminator which scores outputs in the label space $\mathcal{Y}$. It is trained to assign high scores to representative output labels from $p(\yv)$, while assigning low scores to samples from
$p_\theta(\yv)$.
It learns to effectively extract latent constraints that hold over the output space and that are implicitly encoded in the samples from $p(\yv)$. The goal of $p_\theta(\yv|\xv)$ is to produce outputs result in higher scores in the discriminator, satisfying the constraints imposed by $D_\phi$ in the process.

\begin{figure*}[t]
\centering
\subfigure[
Our architecture trains $p_\theta$ by asking it to take in frames and generate trajectories that cannot be discriminated from sample trajectories from a simulator. 
Training $D_\phi$ eliminates the need for hand-engineering constraints.
]{\label{fig:subfig:b}\includegraphics[width=0.4\textwidth]{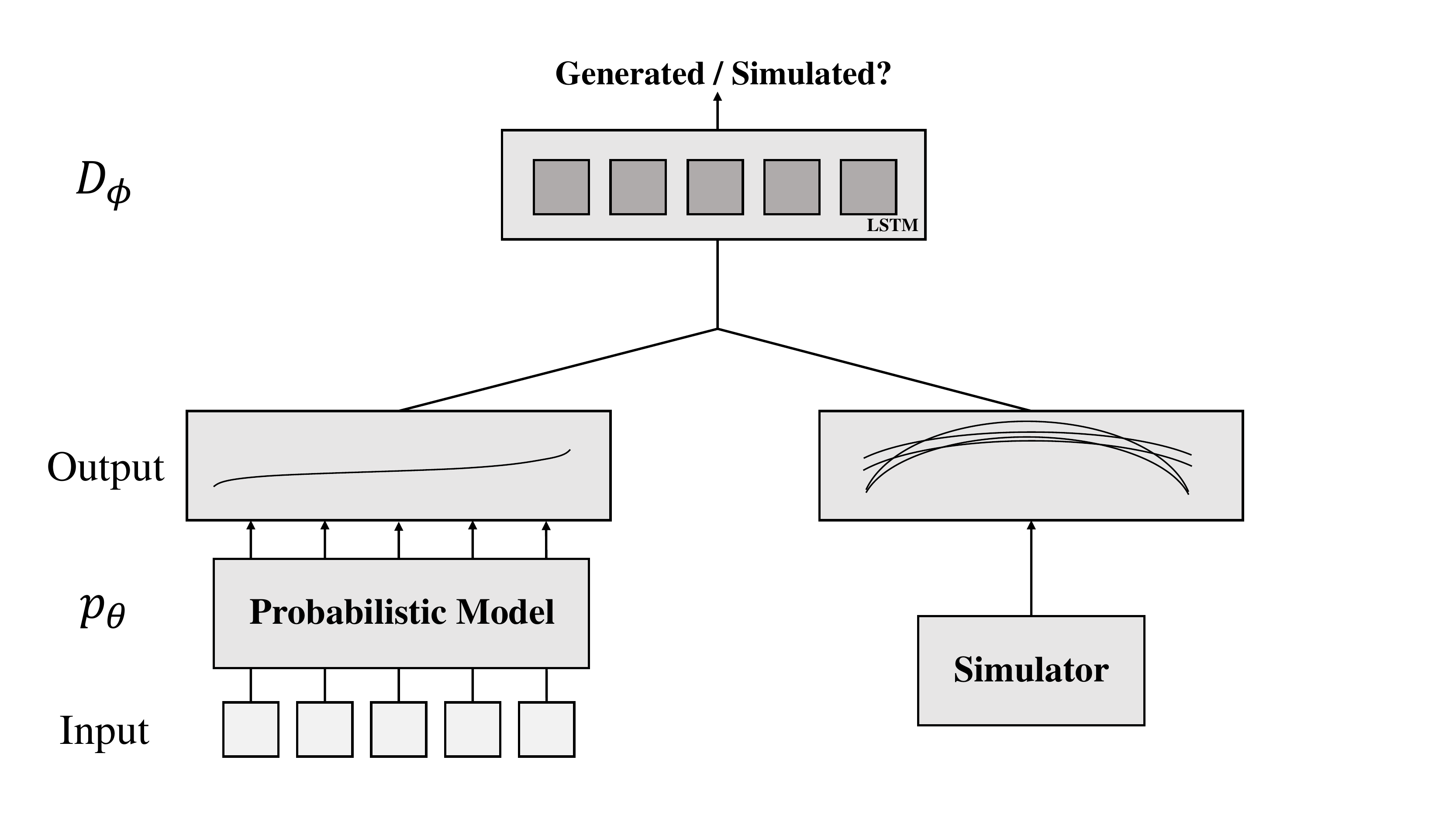}}
~
\subfigure[
Top: frames from the video used in the pendulum experiment. Bottom: the network is trained to predict angles that cannot be distinguished from the simulated dynamics, encouraging it to track the metal ball over time. 
]{\label{fig:subfig:a}\includegraphics[width=0.4\textwidth]{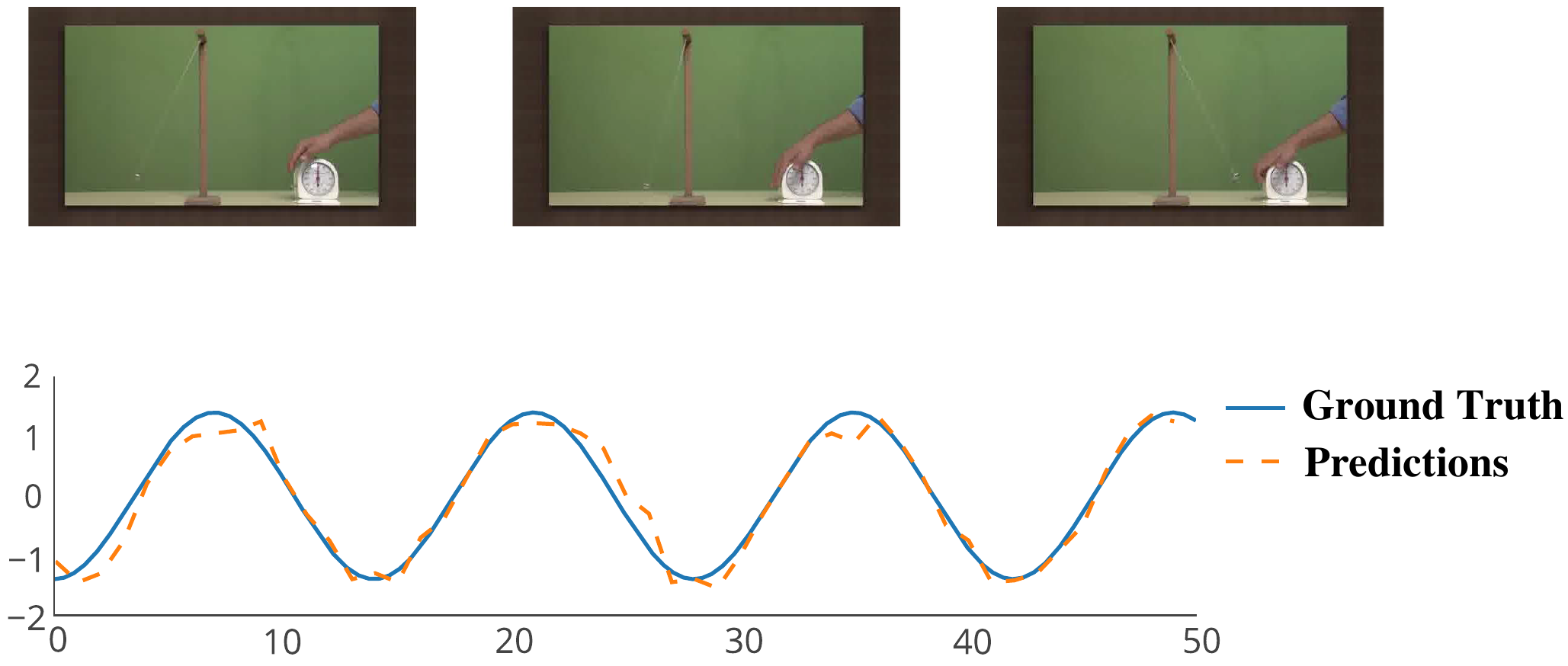}
}
\caption{Architecture and results of the pendulum tracking experiment.
}
\label{fig:pen}
\end{figure*}

For practical reasons, we consider $p_\theta(\yv|\xv)$ to be a Dirac-delta distribution $\delta(\yv-f_\theta(\xv))$, and
thus we refer to the conditional probabilistic model as the mapping $f_\theta(\xv) : \mathcal{X} \to \mathcal{Y}$ in the experiment section for simplicity. 
We train $D_\phi$ and $p_\theta(\yv|\xv)$ for the following objective~\cite{arjovsky2017wasserstein}
\begin{gather}
\min_\theta \max_\phi \mathcal{L^A} \label{eq:probgan}\\
\mathcal{L^A} = \mathbb{E}_{\yv \sim p(\yv)}[D_\phi(\yv)] - \mathbb{E}_{\yv \sim p_\theta(\yv|\xv), \xv \sim p(\xv)}[D_\phi(\yv)] \nonumber
\end{gather}
Assuming 
infinite capacity, Theorem 1 of \cite{goodfellow2014generative} 
shows that at the optimal solution of Eq.~\ref{eq:probgan}, $D_\phi$ cannot distinguish between the given set of labels and those predicted by the model $p_\theta$, suggesting that the latter satisfy the set of constraints defined by $D_\phi$. Unlike in constraint-based learning where a (possibly incomplete) set of constraints is manually specified, convergence in the adversarial setting implies that the label and output distributions match on all possible discriminator projections.
Figure \ref{fig:subfig:b} shows an overview of the adversarial constraint learning framework in the context of trajectory estimation.

\subsection{Constraint Learning via Matching Distributions}
Generative Adversarial Networks (GANs) are a prominent example of implicit probabilistic models~\cite{mohamed2016learning} which are defined through a stochastic sampling procedure instead of an explicitly defined likelihood function. One advantage of implicit generative models is that they can be trained with methods that do not require likelihood evaluations. 

Hence, our approach to learning constraints for structured prediction can also be interpreted as learning an implicit generative model $p_\theta(\yv)$ that matches the empirical label distribution $p(\yv)$. Specifically, our adversarial constraint learning approach optimizes over an approximation to the optimal transport from $p_\theta(\yv)$ to $p(\yv)$~\cite{arjovsky2017wasserstein}; thus our constraint can be implicitly defined as ``$\theta$ minimizes the optimal transport from $p_\theta(\yv)$ to $p(\yv)$''.

\subsection{Semi-Supervised Structured Prediction}

\begin{table}[h]
\centering
\begin{tabular}{c|c|c|c}
Models & $(\xv,\yv)$ & $(\xv,)$ & $(,\yv)$\\
\hline
SL & $\surd$ & &\\
SSL & $\surd$ & $\surd$ & \\
ACL & &$\surd$ &$\surd$\\
SSACL &$\surd$&$\surd$&$\surd$\\
\hline
\end{tabular}
\caption{Settings in different learning paradigms. 
Supervised Learning (SL) requires a dataset with paired $(\xv,\yv)$. 
Semi-Supervised Learning (SSL) 
utilizes additional unlabeled inputs $(\xv,)$. 
Adversarial Constraint Learning (ACL) requires 
inputs $(\xv,)$ and labels $(,\yv)$ but without correspondences between them. 
Semi-Supervised Adversarial Constraint Learning (SSACL) extends ACL by also considering labeled pairs $(\xv,\yv)$.
}
\label{table:models}
\end{table}

Although our framework does not require datasets containing input-label pairs $\Dv \supsetneq \varnothing$, providing it with such data gives rise to a new semi-supervised structured prediction method.

When given a set of labeled examples, we may extend our constraint learning objective (over both labeled and unlabeled data) with a standard classification loss term (over labeled data):
\begin{gather}
\label{eq:advcl}
\mathcal{L^{SS}} = \mathcal{L^A} + \alpha \mathbb{E}_{\xv_i, \yv_i \sim p(\xv,\yv)}[\ell(p_\theta(\yv | \xv_i),\yv_i)]
\end{gather}
where $\mathcal{L^A}$ is the adversarial constraint learning objective defined in Eq.~\ref{eq:probgan}, 
and $\alpha$ is a hyperparameter that balances between fitting to the general (implicit) label distribution (first term) and fitting to the explicit labeled dataset (second term).

Our semi-supervised constraint learning framework is different from traditional semi-supervised learning approaches, as listed in Table \ref{table:models}.
In particular, traditional semi-supervised learning methods assume there is a large source of \emph{inputs} and tend to impose regularization over $\mathcal{X}$, 
 such as through latent variables \cite{kingma2014semi}, through outputs \cite{miyato2017virtual}, or through another network \cite{salimans2016improved}. We consider the case where there exists a source, e.g., a simulator that can provide abundant \emph{samples from the label space that are not matched to particular inputs}, and impose regularization over $\mathcal{Y}$ by exploiting a discriminator that provides an implicit constraint over the predicted $\yv$ values. Therefore, we can also utilize sample labels that are not associated with particular inputs,
instead of merely restricting to standard labeled $(\xv, \yv)$ pairs. Moreover, our method can be easily combined with existing semi-supervised learning approaches \cite{kingma2014semi,li2016max,miyato2017virtual} to further boost performance.

\section{Experimental Results}
\label{experiments}
\label{experiments}

We evaluate the proposed framework on three structured prediction problems. First, we aim to track the angle of a pendulum in a video without labels using supervision provided by a physics-based simulator.
Next, we extend the output space to higher dimensions and perform human pose estimation in a semi-supervised setting. Lastly, we evaluate our approach on multivariate time series prediction, where the goal is to predict future temperature and humidity. 

A label simulator is provided for each experiment in place of hand-written constraints.  
Although explicit constraints for the pendulum case can be written down analytically, we demonstrate that our adversarial framework is capable of learning the constraint from data. 
In the other two experiments, we consider structured prediction settings where the outputs are high dimensional; in these settings, the correct constraints are very complex and hand-engineering them would be difficult. Instead, our model learns these constraints from a small number of samples provided by the simulator.

\subsection{Pendulum Tracking}
For this task, we aim to predict the angle of the pendulum from images in a YouTube video \footnote{https://www.youtube.com/watch?v=02w9lSii\_Hs}, i.e., learn a regression mapping $r_\theta: \mathbb{R}^{h\times{w}\times{3}}\to\mathbb{R}$, 
where $h$ and $w$ are the height and width of the input image. Since the outputs of $r_\theta$ over consecutive frames are constrained by temporal structure (a sine wave in this case), 
we concatenate consecutive outputs of $r_\theta$ and form a high dimensional trajectory, thus defining $f_\theta([\xv_1, \xv_2, \cdots, \xv_n]) = [r_\theta(\xv_1), r_\theta(\xv_2), \cdots, r_\theta(\xv_n)]$. Critically, $r_\theta$ must make a separate prediction for each image, preventing $f_\theta$ from simply memorizing the output structure.
Unlike previous methods~\cite{stewart2017label}, no explicit formulas are provided for supervision, and the (implicit) constraints are learned through the discriminator $D_\phi$ using samples provided by the physics simulator.

\paragraph{Training Details}
The video contains a total of 170 images, and we hold out $34$ images for evaluation. We manually observe that the pendulum completes one full oscillation approximately every 12 frames. Based on this observation, we write a simulator of these dynamics with a simple harmonic oscillator having a fixed amplitude and random sample period of 10 to 14 frames. 
$D_\phi$ is trained to distinguish between the output of $r_\theta$ across $n=5$ continuous images and a random trajectory sampled from the simulator. We implement $r_\theta$ as a 5 layer convolutional neural network with ReLU nonlinearities, and $D_\phi$ as a 5-cell LSTM. We use $\alpha=10$ in Eq.~\ref{eq:advcl}, and the same training procedure and hyperparameters as~\cite{gulrajani2017improved} across our experiments.

\paragraph{Evaluation}
We manually label the horizontal position of the ball of the pendulum for each frame \textit{in the test set}, and measure the correlation between the predicted positions and the ground truth labels. Since the same $r_\theta$ is applied to each input frame independently, $f_\theta$ cannot just memorize valid (i.e. simple harmonic) trajectory sequences and produce them while ignoring inputs. The model must learn to track the pendulum in order to fool the discriminator and subsequently achieve a high correlation on the test set.

Our adversarial constraint learning approach achieves a correlation of $96.3\%$, whereas training with hand-crafted constraints achieves a marginally higher correlation of $96.6\%$. Both approaches are trained without labels.
Example predictions on the test data are shown in Figure~\ref{fig:pen}. 
This real-world experiment demonstrates the effectiveness of constraint-based learning in the absence of labels, and suggests that using learned constraints from data is almost as effective as using ideal hand-crafted constraints.

\begin{figure*}
\centering
\includegraphics[width=0.75\textwidth]{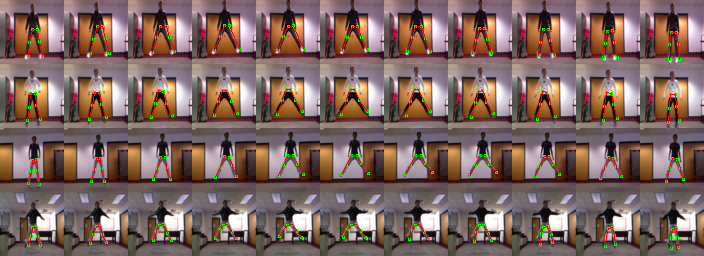}
\caption{Pose estimation using the proposed semi-supervised adversarial constraint learning approach. $r_\theta$ takes in single image and outputs the 2-D location of 6 joints (in green). Lines (in red) are added automatically. The images show results across 4 test groups (horizontal strips) when only 3 out of 28 training groups were directly labeled.}
\label{fig:pose}
\end{figure*}

\subsection{Pose Estimation}
In this experiment, we evaluate the proposed model on pose estimation, which has a significantly larger output space. We aim to learn a regression network $r_\theta: \mathbb{R}^{h\times{w}\times{3}}\to\mathbb{R}^{k\times{2}}$, where 
$k$ denotes the number of joints to detect, and each joint has $2$ coordinates. As in the pendulum tracking experiment, $r_\theta$ is mapped across several frames to produce a trajectory $f_\theta([\xv_1, \xv_2, \cdots, \xv_n]) = [r_\theta(\xv_1), r_\theta(\xv_2), \cdots, r_\theta(\xv_n)]$ that is indistinguishable from samples provided by the simulator.

We evaluate the model with videos and joint trajectories from the CMU multi-modal action database (MAD)~\cite{huang2014sequential}. MAD contains videos of 20 subjects performing a sequence of actions in each video. We extract frames from subjects performing the ``Jump and Side-Kick'' action and train $r_\theta$ to detect the location of the left/right hip/knee/foot ($k=6$) in each frame. The processed dataset contains 35 groups (549 valid frames in total).

\paragraph{Training Details}
We divide the 35 groups of motion data into training and testing sets of 28 groups and 7 groups, respectively, where direct labels will be provided for a subset of the 28 training groups. Each group contains 14 to 17 frames, and we train on randomly selected contiguous intervals of length $n=5$. Using the metric of PCK@0.1~\cite{yang2013articulated} for evaluation, a prediction is considered correct if it lies within $\beta\max(h,w)$ pixels from the true location, where $h$ and $w$ denote the height and width of the subject's body. 
We evaluate with $\beta=0.1$. 

We first design a simulator of valid labels (joint positions) based on known kinematics of skeletons. Specifically, the anatomical shape of the subject's legs approximately forms an expanding isosceles trapezoid when they jump and side kick. We simulate a large range of trapezoidal motions capturing these trajectories, which requires much less effort than hand engineering precise mathematical formulas to express explicit constraints.
$r_\theta$ takes a single image as input and produces a $12$ dimensional vector, representing the location of $k=6$ joints.
Critically, as in the pendulum experiment, $r_\theta$ is applied to each frame independently, and has \textit{no knowledge} of the neighboring frames. The outputs of $r_\theta$ are concatenated and passed to the discriminator $D_\phi$ for training. 

\paragraph{Evaluation}
We construct 50 random train/test splits of the dataset and report the averaged PCK@0.1 scores for evaluation.
The results are summarized in Table \ref{table:pose} and Figure \ref{fig:pose}, where we compare three forms of learning when labels are only available for $i \leq 28$ of the training groups: 
\begin{itemize}
\item{``L(i)'': vanilla supervised learning on labeled groups}
\item{``L(i)+VAT'': a baseline form of semi-supervised learning leveraging virtual adversarial training on unlabeled groups (VAT,~\cite{miyato2017virtual})}
\item{``L(i)+ADV'': semi-supervised learning with adversarial constraint learning (Eq.~\ref{eq:advcl})}
\end{itemize}
When no labels are provided (``L(0)+ADV''; i.e., optimizing just $\mathcal{L^A}$), $r_\theta$ is able to find the correct ``shape'' of the joints for each frame, but the predictions are biased. Since the subjects are not strictly acting in the center of the image, a constant minor shift ($\Delta{x},\Delta{y}$) for all predicted joint locations still meets the requirements imposed by $D_\phi$, which encodes the structure of the label space. This problem is addressed when providing even a very small (``i=1'') number of labeled training groups and using the semi-supervised objective $\mathcal{L^{SS}}$. Availability of labels fixes the constant bias and we note that using adversarial training produces a massive (25-30\%) boost over both the supervised and ``VAT'' baselines when only 1 group of labeled data is available.

With only 3 groups of labeled data (``L(3)+ADV''), adversarial constraint learning achieves a comparable performance to standard supervised learning with 7 groups of labeled inputs (``L(7)''). Adversarial constraint learning ``L(i)+ADV'' consistently outperforms the virtual adversarial training ``L(i)+VAT'' baseline for different values of $i$. When further combined with VAT regularization in the $\mc{L^{SS}}$ objective, our method achieves slightly better performance. 

The strong performance of our model over baselines on the pose estimation task with a few or no labels demonstrates that constraint learning can work well over high-dimensional outputs when using our proposed adversarial framework. Designing precise constraints in high-dimensional spaces is often tedious, error-prone, and restricted to one particular domain. Our method avoids these downsides by learning these constraints implicitly through data generated from a simulator, even though the simulator can be a noisy (or even slightly biased) description of the true label distribution.

\begin{table*}[tbp]
\centering
\begin{tabular}{c|c|c|c|c|c|c}
PCK@0.1(\%) & Left Hip & Left Knee & Left Foot & Right Hip & Right Knee & Right Foot\\
\hline
L(0)+ADV &0.6813&0.7326&0.6047&0.6669&0.6729&0.5834\\
\hline
L(1) & 0.5453&0.5728&0.5464&0.5360&0.4983&0.4362\\
L(1)+VAT &0.5795&0.6086&0.5797&0.5608&0.5016&0.4571\\
L(1)+ADV & \textbf{0.8529} & \textbf{0.8510} & \textbf{0.8151} & \textbf{0.8482} & \textbf{0.8531} & \textbf{0.7394}\\
\hline
L(3) & 0.8275&0.7937&0.6716&0.8092&0.7529&0.6196\\
L(3)+VAT &0.8334&0.7866&0.7082&0.8281&0.7760&0.6420\\
L(3)+ADV & \textbf{0.8760} & \textbf{0.9097} & \textbf{0.8328} & \textbf{0.8601} & \textbf{0.8746} & \textbf{0.7549}\\
\hline
L(5) &0.8603&0.8483&0.7493&0.8309&0.8267&0.6626\\
L(5)+VAT &0.8750&0.8764&0.7411&0.8471&0.8398&0.6596\\
L(5)+ADV & \textbf{0.9022} & \textbf{0.9160} & \textbf{0.8581} & \textbf{0.9192} & \textbf{0.8706} & \textbf{0.7894}\\
\hline
L(7) &0.9088&0.8639&0.8217&0.8887&0.8387&0.7338\\
L(7)+VAT &0.9201&0.8665& \textbf{0.8436} &0.9074&0.8312&0.7526\\
L(7)+ADV & \textbf{0.9469} & \textbf{0.9347} &0.8418& \textbf{0.9367} & \textbf{0.8988} & \textbf{0.8161}\\
\hline
L(ALL) &0.9622&0.9633&0.9290&0.9464&0.9133& \textbf{0.8936}\\
L(ALL)+ADV & \textbf{0.9758} & \textbf{0.9882} & \textbf{0.9627} & \textbf{0.9708} & \textbf{0.9522} &0.8740\\
\hline
\end{tabular}
\caption{PCK@0.1 results on MAD. ``L(i)'' indicates supervised learning (SL) where labeled data is provided for only $i$ out of 28 groups in the training set. ``L(i)+VAT'' indicates SL with additional optimization over unlabeled groups using virtual adversarial training \protect\cite{miyato2017virtual} (SSL). ``L(i)+ADV'' indicates SL with additional optimization over the entire training set with the ACL objective. Our approach outperforms the baselines, especially when very few labels are available. 
}
\label{table:pose}
\end{table*}

\subsection{Time Series Prediction}
Lastly, we validate our model on another structured prediction problem: multi-step multivariate time series prediction. In this task, we aim to learn a mapping $f_\theta: \mathbb{R}^{t\times{m}}\to\mathbb{R}^{k\times{m}}$.
Given $t$ consecutive values of a series, $(\yv_1, \yv_2, \cdots, \yv_{t})$, we aim to predict the following $k$ values, $(\yv_{t+1}, \cdots, \yv_{t+k})$, where each $\yv$ has $m$ variables. 
In this task, $D_\phi$ learns the constraint that both holds across variables and time by distinguishing the output of $f_\theta$ from real label samples.

\begin{figure}
\centering
\includegraphics[width=0.45\textwidth]{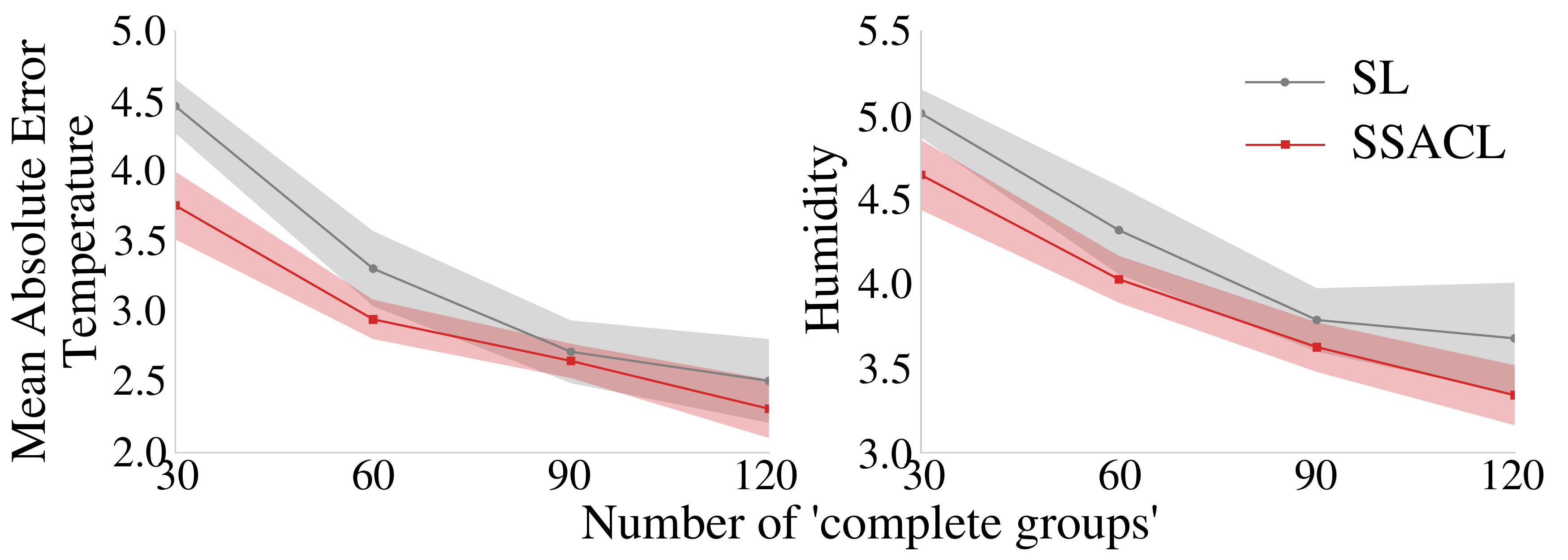}
\caption{Mean absolute error of the predictions on temperature (top) and humidity (bottom) during training (left) and testing (right). Our method SSACL (trained on $\mathcal{L}^{SS}$ objective) consistently outperforms SL (supervised learning) on the test set with different numbers of ``complete groups'' used in training.
}
\label{fig:time}
\end{figure}

\paragraph{Training Details}

We conduct experiments on the SML2010 Dataset~\cite{zamora2014line}, which contains humidity and temperature data of indoor and outdoor environments over 40 days at 15 minute intervals. 

We hold out 8 consecutive days for testing and leave the rest for training. From the train and test set, we sample 480 and 120 groups of time series data respectively, each having length of 28 hours, and smooth each group into 7 data points at 4-hour intervals.
Each group uses the first $t=5$ data points as input, and leaves the final $k=2$ values as targets for prediction, with each data point having $m=4$ variables representing the indoor/outdoor temperature/humidity. We measure the mean absolute error (MAE) on the test set. 

We further explore the setting when not all groups in the training set are ``complete''; for example, in some groups we may only have temperature information. This is reasonable in real-world scenarios where  sensors fail to work properly from time to time. Hence, we use ``complete groups'' to denote groups with full information, and ``incomplete groups'' to denote groups with only temperature information.
Without humidity records, we could not perform supervised learning on these ``incomplete groups'', since the input requires all $m=4$ values. 
Under the context of adversarial constraint learning, however, such data can facilitate learning constraints over the temperature series. In this task, the simulator is designed to produce humidity samples from only the ``complete groups'' and temperature samples from all the groups in the training set. Both $f_\theta$ and $D_\phi$ are 4-layer MLPs with 64 neurons per layer.

\paragraph{Evaluation}
We display quantitative results in Figure \ref{fig:time}. The supervised baseline (trained only with labels) achieves lower training error but results in higher test error. Our model effectively avoids overfitting to the small portion of labeled data and consistently outperforms the baseline, achieving a MAE of 1.933 and 3.042 on the predictions of temperature and humidity when all groups are ``completely'' labeled.

\section{Conclusion}
We have proposed adversarial constraint learning, a new framework for structured prediction that replaces hand-crafted, domain specific constraints with implicit, domain agnostic ones learned through adversarial methods. 
Experimental results on multiple structured prediction tasks demonstrate that adversarial constraint learning works across many real-world applications with limited data, and fits naturally into semi-supervised structured prediction problems. Our success with matching distributions of labeled and unlabeled model outputs motivates future work exploring analogous opportunities for adversarially matching labeled and unlabeled distributions of learned intermediate representations.

\section*{Acknowledgments}
This work was supported by a grant from the SAIL-Toyota Center for AI Research, TRI, Siemens, ONR, NSF grants \#1651565, \#1522054, \#1733686, and a Hellman Faculty Fellowship.

\bibliographystyle{named}
\bibliography{ijcai18}

\end{document}